%% file: interspeech2021-kglm.tex
\documentclass[a4paper]{article}

\usepackage{titlesec}
\usepackage[numbers,square]{natbib} 
\usepackage{INTERSPEECH2020}
\usepackage{amsmath}
\usepackage{adjustbox}
\usepackage[english]{babel}
\usepackage{booktabs}
\usepackage{hyperref}
\usepackage[inline]{enumitem}
\usepackage{listings}
\usepackage{graphicx}
\usepackage[utf8x]{inputenc}
\usepackage{microtype}
\usepackage{multirow}
\usepackage{numprint}
\usepackage{newunicodechar}
\usepackage[nodisplayskipstretch]{setspace}
\usepackage[dvipsnames]{xcolor}
\usepackage[font=small,skip=0pt,belowskip=0pt]{caption}
\usepackage{algorithm2e}
\usepackage{xstring}
\usepackage{xifthen}

\titlespacing\section{0pt}{4pt plus 2pt minus 2pt}{0pt plus 2pt minus 2pt}
\titlespacing\subsection{0pt}{2pt plus 1pt minus 1pt}{0pt plus 1pt minus 1pt}

\colorlet{punct}{red!60!black}
\definecolor{background}{HTML}{EEEEEE}
\definecolor{delim}{RGB}{20,105,176}
\colorlet{numb}{magenta!60!black}

\lstdefinelanguage{json}{
    basicstyle=\scriptsize\ttfamily,
    stepnumber=1,
    numbersep=8pt,
    showstringspaces=false,
    breaklines=false,
    frame=lines,
    backgroundcolor=\color{background},
    literate=
     *{0}{{{\color{numb}0}}}{1}
      {1}{{{\color{numb}1}}}{1}
      {2}{{{\color{numb}2}}}{1}
      {3}{{{\color{numb}3}}}{1}
      {4}{{{\color{numb}4}}}{1}
      {5}{{{\color{numb}5}}}{1}
      {6}{{{\color{numb}6}}}{1}
      {7}{{{\color{numb}7}}}{1}
      {8}{{{\color{numb}8}}}{1}
      {9}{{{\color{numb}9}}}{1}
      {:}{{{\color{punct}{:}}}}{1}
      {,}{{{\color{punct}{,}}}}{1}
      {\{}{{{\color{delim}{\{}}}}{1}
      {\}}{{{\color{delim}{\}}}}}{1}
      {[}{{{\color{delim}{[}}}}{1}
      {]}{{{\color{delim}{]}}}}{1},
}

\addto\extrasenglish{%
}

\newcommand{\tb}{\textbf}

\newunicodechar{〈}{\langle}
\newunicodechar{〉}{\rangle}

\newcommand{\Slot}[1]{\${}#1}
\newcommand{\EntityEntityRelationship}{entity-entity}
\newcommand{\EntityTypeEntityRelationship}{entity type~--~entity}

\title{A Discriminative Entity-Aware Language Model\\for Virtual Assistants}

\name{%
\begin{tabular}{c}%
Mandana Saebi\thanks{$^1$The work described in this paper was performed while the first author was an intern at Apple.}$^{1}$ \quad Ernest Pusateri$^2$ \quad Aaksha Meghawat$^2$ \quad Christophe Van Gysel$^2$%
\end{tabular}}
\address{
    \textsuperscript{$^1$}University of Notre Dame, Notre Dame, IN, USA,\\
    \textsuperscript{$^2$}Apple, Cupertino, CA, USA
}
\email{
    \texttt{\href{mailto:msaebi@nd.edu}{msaebi@nd.edu}},
    \texttt{\{\href{mailto:epusateri@apple.com}{epusateri},
              \href{mailto:ameghawat@apple.com}{ameghawat},
              \href{mailto:cvangysel@apple.com}{cvangysel}\}@apple.com}}

\begin{document}
\maketitle

\begin{abstract}
High-quality automatic speech recognition (ASR) is essential for virtual assistants (VAs) to work well. However, ASR often performs poorly on VA requests containing named entities. In this work, we start from the observation that many ASR errors on named entities are inconsistent with real-world knowledge.  We extend previous discriminative n-gram language modeling approaches to incorporate real-world knowledge from a Knowledge Graph (KG), using features that capture entity type-entity and entity-entity relationships. We apply our model through an efficient lattice rescoring process, achieving relative sentence error rate reductions of more than 25\% on some synthesized test sets covering less popular entities, with minimal degradation on a uniformly sampled VA test set.

\end{abstract}

\noindent\textbf{Index Terms}: Speech Recognition, Language Modeling, Semantic knowledge, Spoken Entities 

\input{01-introduction}

\input{02-approach}

\input{03-related_work}
\input{04-experiments}
\input{05-conclusions}
\bibliographystyle{IEEEtranN}
\bibliography{interspeech2021-kglm}

\end{document}

%% file: 01-introduction.tex
\section{Introduction}
\label{sec:intro}

Digital virtual assistants (VAs) continue to gain popularity and reach \cite{Juniper2019popularity}, propelled by their ability to respond to a wide range of requests made with natural human speech. High-quality automatic speech recognition (ASR) is essential for these systems to work well. However, an important subset of requests, those containing named entities, still presents a significant challenge to ASR \citep{VanGysel2020entities}. This is primarily because it is difficult to build a language model (LM) that accurately models entities that are less popular or only recently popular, as they show up rarely or not at all in training data.  See Figure 1 for an example of the problem we describe. Given the VA request ``\textit{Play Canyon Moon by Harry Styles}'', ASR returns a top choice hypothesis which substitutes an acoustically similar phrase ("Can you Moon") for the correct entity name (``Canyon Moon".)

In this work, we start from the observation that many ASR named entity errors are inconsistent with real-world knowledge.  Returning to Figure 1, knowing the songs performed by Harry Styles would make it obvious that the fourth hypothesis was the correct one.  Building on this intuition, our approach uses real-world knowledge in the form of a knowledge graph (KG) to improve ASR performance on named entities.

We bring a KG to bear through a discriminative entity-aware language model (DEAL) which builds on an n-gram discriminative language modeling approach introduced in \cite{roark2004discriminative}.  In that work, a log-linear model consumes features including a base model score and word n-gram indicators. The model is trained on n-best lists or lattices generated from a base system to minimize a discriminative loss function.  We extend that approach by defining n-gram indicator features that fire when a word sequence within an ASR hypothesis is consistent with a KG.  We design our features to capture \EntityTypeEntityRelationship{} relationships (e.g. ``Canyon Moon'' is a song) as well as relationships between entities (e.g. Canyon Moon is a song by Harry Styles.)  We apply the model using an efficient lattice rescoring process.  By applying this approach, we are able to achieve large improvements on targeted test sets covering unpopular entities with minimal degradation on a uniformly sampled VA test set.

Our main contributions are: %
\begin{enumerate*}[label=(\arabic*),noitemsep,topsep=0pt,parsep=0pt,partopsep=0pt,itemjoin=\quad]
\item a discriminative language modeling approach for incorporating knowledge from a KG into an ASR system,
\item a performance analysis of different feature sets, and
\item an efficient algorithm to apply the approach via lattice rescoring.
\end{enumerate*}

\begin{figure}
    \centering
    \includegraphics[width=1\linewidth]{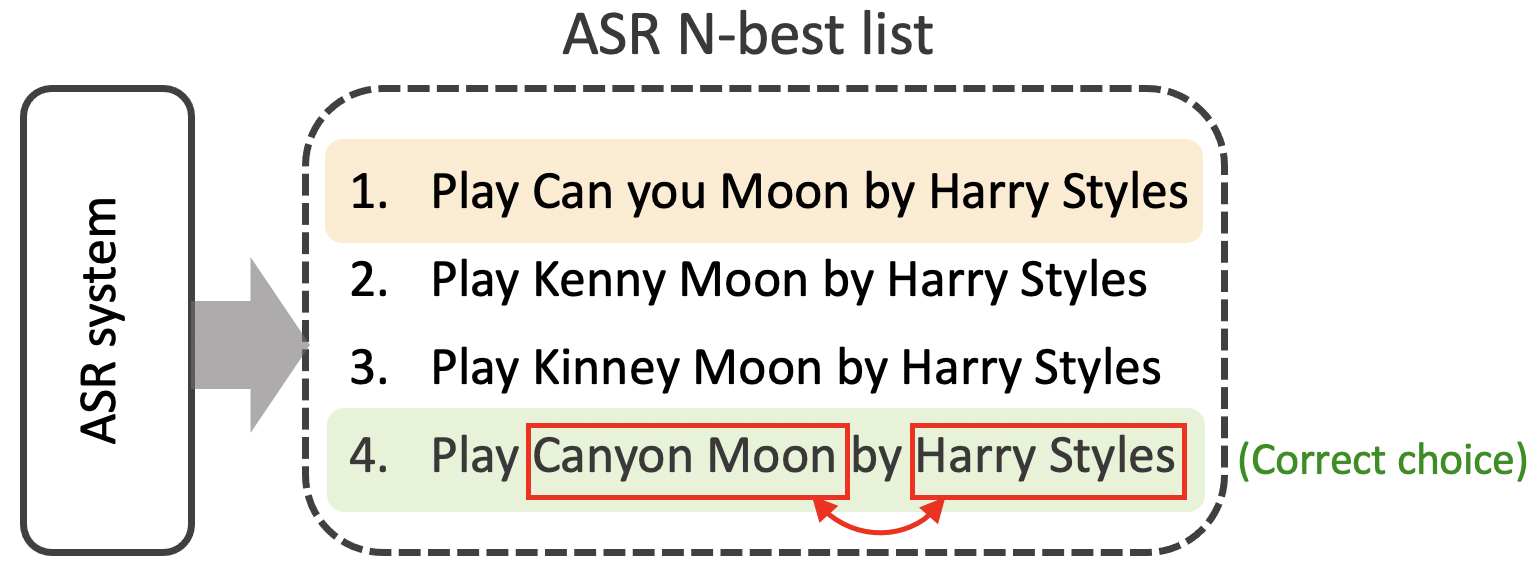}
    \caption{Example of a synthesized VA utterance and ASR hypothesis. The ASR top choice can be improved using knowledge of the relationship between \textit{Canyon Moon} and \textit{Harry Styles}.}
    \label{fig:my_label}
\label{fig:intro_example}
\end{figure}

%% file: 02-approach.tex
\section{Approach}

\begin{table*}[t]%
\centering\noindent%
\begin{tabular}{m{0.37\textwidth}m{0.57\textwidth}}
    \hspace{0.5cm}
    \begin{tabular}{ccc}
        id & n-gram & weight \\
        \midrule
        f1 & play \Slot{title} by & 1.2 \\
        f2 & play \Slot{artist} & 0.8 \\
        f3 & to \Slot{city} \Slot{state$|$city} & -0.4 \\
    \end{tabular}%
    &
    \hspace{0.2cm}%
    \includegraphics[height=2.25cm]{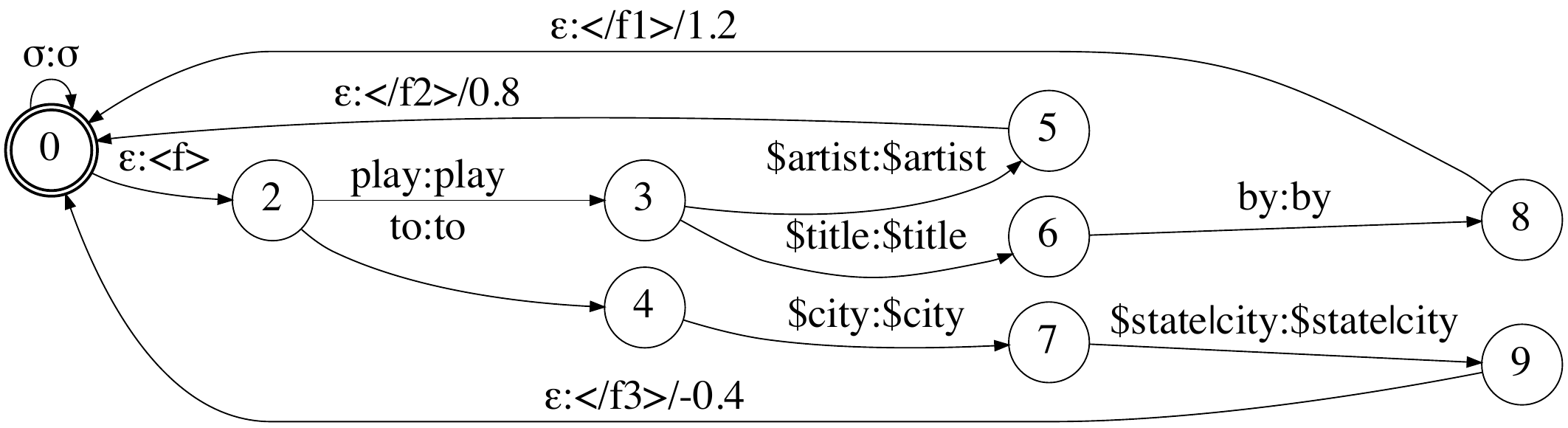}%
\end{tabular}%
\captionof{figure}{An example of features, their learned weights and the $R_{top}$ built from them using \autoref{alg:rtop}.  \Slot{state$|$city} represents a non-terminal covering states with a contains relation to the realization of the preceding \Slot{city} non-terminal. The $\sigma$ symbol in $R_{top}$ matches all input symbols and reproduces them on the output.}
\label{fig:lr}
\end{table*}

Following \cite{roark2004discriminative}, the model underlying our approach takes the form of a set of features and weights. The first feature is defined as the score from a base system. All other features are specified by n-grams.  N-gram features fire when the corresponding n-gram matches a word sequence within a hypothesis.  At test time, the features and weights are used to rescore n-best lists or lattices generated from the base system.  The score given to hypothesis $h$ in utterance $u$ is shown in \autoref{eq:score}, where $w_f$ is the learned weight for feature $f$, and $x_{u,h,f}$ is the value of feature $f$, for hypothesis $h$, utterance $u$.  $x_{u,h,0}$ always takes the value of the combined LM and acoustic model score from the base system.  At training time, n-best lists are generated for a training set using the base system, and the weights are learned to minimize a discriminative loss function.
\begin{equation}
s(u,h) = \sum_{f=0}^{F}{w_{f}x_{u,h,f}}
\label{eq:score}
\end{equation}
In the following subsections, we describe how we extended \cite{roark2004discriminative} to incorporate information from a KG via enhanced features, and how we were able to use our approach to efficiently do lattice rescoring.

\subsection{Features}

\label{features}

\begin{figure}[t]
\centering
\begin{lstlisting}[language=json]
"12345": {
    "names": {"Canyon Moon": {"word count": 2}},
    "types": {"music title": {"popularity": 0.0025}},
    "relationships": [{
      "relation": "performed by",
      "entity id": "67890",
      "popularity": 0.0021}]}
"67890": {
    "names": {
        "Harry Styles" : {"word count": 2},
        "Harry Edward Styles": {"word count": 3}
    },
    "types": {"music artist": {"popularity": 0.5}},
    "relationships": [{
       "relation": "performed",
       "entity id": 12345,
       "popularity": 0.1}]}
\end{lstlisting}
\caption{Example of two related entities in our KG.}
\label{fig:KG}
\end{figure}

Our features reference data stored in our KG, so we will start with a description of the KG's structure. Each node in our KG is an \textit{entity}, which represents a real-world object. An entity has \textit{names}, which are word sequences used to denote it, \textit{types}, which are broad semantic classes to which the entities belong and \textit{relationships}, which represent connections to other entities. A name has a \textit{word count}, which is the number of words in the entity. A relationship has an \textit{entity id}, which specifies the related entity, a \textit{relation} which defines the nature of the relationship, and a \textit{popularity}, which indicates the frequency with which the entity and its related entity appear together. \autoref{fig:KG} shows an example of two related entities.

Our features are specified as n-grams comprised of words and at least one \textit{non-terminal}. A non-terminal represents the union of the entity names sharing an entity type, possibly with conditions. Non-terminals can be conditioned on entity name word count or entity popularity, as well as entities whose names were previously seen in a hypothesis. A feature takes the value 1 for a word sequence in a hypothesis when the word sequence matches a valid realization of the n-gram. A realization is a word sequence generated by replacing each non-terminal with a corresponding entity name. If the feature does not match a valid realization, the value is set to 0.

To take into account entity popularity, for each entity type, we create three non-terminals, the first covering the most popular (head) entities, the second adding moderately popular (torso) entities, and the last adding the least popular (tail) entities. We also create three non-terminals for each entity type based on entity name word count, covering entity names with one or more words, two or more words, and three or more words. The entity lists corresponding to each conditional non-terminal intentionally overlap to reduce the number of features with few training examples.  By using entity popularity and word count, we hoped to expand the space of trade-offs the model could make. For instance, we hypothesized that the model can be more aggressive in boosting hypotheses for popular entities versus less popular ones as the risk of causing an error by doing so would be lower. Similarly, for entity name word count, we surmised that the model could more aggressively reward hypotheses with longer entity names, as it would be less likely for these to show up incorrectly in an ASR lattice.

To choose feature n-grams, we first derive text templates with frequency estimates. These templates consist of entity type non-terminals (without conditions) and surrounding word context and cover the most common requests where the targeted entity types or related entity type pairs appeared. For instance, the most common template for the city and state entity type pair was ``Directions to \Slot{city} \Slot{state}”. From these templates, we extract all 3-grams containing a non-terminal. In order to better model entity relationships, we also include all 4-grams that begin and end with a non-terminal. We then factor the resulting n-grams into multiple n-grams with conditional non-terminals. Examples of feature n-grams can be found in Figure \ref{fig:lr}.

\subsection{Lattice Rescoring}

While it is straightforward to use our approach to do n-best rescoring, we also developed a way of using it to rescore lattices. We prefer rescoring lattices to n-best lists because lattices can represent a larger set of hypotheses more efficiently, and an efficient lattice rescoring scheme can avoid the redundant computations that occur with n-best rescoring due to the large overlap between nearby hypotheses.

In the approach described in \cite{roark2004discriminative}, a lattice, represented as a weighted finite-state automaton (WFSA) \cite{mohri2002weighted}, is rescored by composing it with a static, deterministic WFSA representing the learned model. That approach does not work for us here, as it would require expanding the non-terminals in all of the features for all of the corresponding entities in the KG. This would produce prohibitively large WFSAs for many use cases and also require that the resulting rescoring WFSA be rebuilt whenever the KG changed. To overcome these difficulties, we developed an approach that constructs a deterministic WFSA on-demand.

Our deterministic rescoring WFSA, $R_{det}$, is built by determinizing a weighted finite-state transducer (WFST), $R_{kg}$, on-demand. $R_{kg}$ is constructed on-demand from another WFST, $R_{top}$, and the KG. $R_{top}$ is built offline and can be thought of as a top-level feature tagging grammar. It is built from the model features and weights according to \autoref{alg:rtop}. An example $R_{top}$ is shown in Figure \ref{fig:lr}.
$R_{kg}$ is built by traversing $R_{top}$ and replacing arcs labeled with non-terminals with WFSAs constructed from their corresponding entity names from the KG. In order to support non-terminals whose realizations depend on previously seen entities, non-terminal/realization pairs are stored as part of the dynamically created $R_{kg}$ states. Non-terminal realizations are propagated until a feature end tag is reached. The result of composing a lattice with $R_{kg}$ would be a WFST with two important properties: %
\begin{enumerate*}[label=(\arabic*),noitemsep,topsep=0pt,parsep=0pt,partopsep=0pt,itemjoin=\quad]
\item exactly one sub-path would exist for any lattice sub-path and n-gram feature realization spanning that sub-path, and %
\item all arcs except those with feature end tags would have 0 weights.
\end{enumerate*}

\begin{algorithm}
\SetKwFunction{FEncode}{Encode}
\SetKwFunction{FDecode}{Decode}
\SetKwFunction{FDeterminize}{Determinize}
\DontPrintSemicolon
\KwIn{$features$, a list of $id$, $ngram$, $weight$ triples}
Initialize $R_{top}$ with start and final state, $s_{top}$ \;
Add arc from $s_{top}$ to $s_{top}$ with symbols $\sigma$:$\sigma$ \;
\ForEach{$id$, $ngram$, $weight$ in $features$}{
    Construct a WFSA that accepts $ngram$, with start state $s_f$, and end state $e_f$ \;
    Add arc from $s_{top}$ to $s_f$ with symbols $\langle$f$\rangle$:$\langle$f$\rangle$ \;
    Add arc from $e_f$ to $s_{top}$ with symbols $\epsilon\colon\langle id \rangle$ and weight $weight$ \;
}
$R_{top} \gets \FDecode(\FDeterminize(\FEncode(R_{top})))$ \;
\caption[Algorithm for constructing $R_{top}$ from feature n-grams and weights. Encode encodes arc input symbols, output symbols and weights in the arc symbols. Decode reverses that operation.]{Algorithm for constructing $R_{top}$ from feature n-grams and weights. $\FEncode$ encodes arc input symbols, output symbols and weights in the arc symbols. $\FDecode$ reverses that operation.}
 \label{alg:rtop}
\end{algorithm}

$R_{det}$ is built by determinizing $R_{kg}$ on-demand.  Note that this determinization operation is not equivalent to conventional determinization on the tropical or log semiring, as we want to apply the same operation, addition, to feature weights on consecutive arcs as well as to paths that converge to the same state.  However, because we have ensured the properties above, the on-demand determinization process is straightforward.  When the composition process requests an arc weight and end state from $R_{det}$ for a start state, $s$  and symbol $i$, the on-demand generation process will find all states in $R_{kg}$ corresponding to $s$ and find the set of arcs, $A$, with input symbol $i$ exiting those states. A new $R_{det}$ arc will be created with a weight equal to the sum of the weights on $A$. The end state will be a new $R_{det}$ state associated with all of the $R_{kg}$ end states on the arcs in $A$, and the input and output symbols will be $i$.

%% file: 03-related_work.tex
\section{Related Work}
\label{ssec:related}
Our approach was particularly inspired by \cite{logan2019barack}, which presents a generative, neural network LM that incorporates a KG. A similar approach is presented in \cite{hayashi2020latent}.  Roughly speaking, one can view our work as a non-neural analog to those approaches with a discriminative optimization criterion. 

While they do not use a KG as their knowledge source, \cite{hall2015composition}, \cite{scheiner2016voice}, \cite{aleksic2015bringing} and \cite{velikovich2018semantic} share many similarities with our work.  The models in all of those works can be viewed as log-linear models which consume n-gram features.  Further, WFSTs are used to encode the models and WFST composition is used to apply them, and all of those works focus on improving ASR for virtual assistants.   However, those approaches differ from ours in that model weights are determined by heuristics, sometimes in combination with context-specific n-gram LMs, rather than through discriminative training. The approaches in \cite{hall2015composition}, \cite{scheiner2016voice}, \cite{aleksic2015bringing} additionally differ in that the models are applied early in the recognition process rather than during lattice rescoring (an advantage, all else equal.)  The approach in \cite{velikovich2018semantic} is especially similar to our work in that their analog to n-gram features can contain non-terminals.   However, in \cite{velikovich2018semantic}, lattices are semantically tagged to determine the locations of the non-terminals before being rescored, and their n-gram features do not capture entity-entity relationships.

%% file: 04-experiments.tex
\section{Experiments}
\label{sec:experiments}
\begin{table*}[t]
\renewcommand{\arraystretch}{0.65}%
\setlength{\tabcolsep}{5pt}%
\small%
\centering%
\caption{Experiment results. \# entities indicates the number of unique entities in each dataset and \# relations indicate the number of unique \EntityEntityRelationship{} relations. Bold numbers indicate the lowest sentence error rate (SER) w.r.t each baseline.\label{tab:results}}%
\resizebox{1.0\textwidth}{!}{%
\newcommand{\MultiLineCell}[1]{%
\parbox{2cm}{%
\setlength{\baselineskip}{0pt}%
{\footnotesize #1}%
}%
}%
\newcommand{\PrintDecimal}[1]{%
\StrLen{#1}[\inputStrLen]%
\ifthenelse{\inputStrLen = 4}{\phantom{00}#1}{\ifthenelse{\inputStrLen = 5}{\phantom{0}#1}{#1}}%
}%
\begin{tabular}{lc|cc|cccccc|c|ccc|c}
\toprule
Data  &  Tier                   & 
\rotatebox[origin=c]{-45}{\# entities} & 
\rotatebox[origin=c]{-45}{\# relations} & 
\rotatebox[origin=c]{-45}{Baseline-1}  &  \rotatebox[origin=c]{-45}{DEAL1}   &  \rotatebox[origin=c]{-45}{DEAL11-r}  &  \rotatebox[origin=c]{-45}{DEAL11-rc}  &  \rotatebox[origin=c]{-45}{DEAL1-rp}  &  \rotatebox[origin=c]{-45}{DEAL1-rpc}  &  \rotatebox[origin=c]{-45}{Oracle-1}  &  \rotatebox[origin=c]{-45}{Baseline-2}  &  \rotatebox[origin=c]{-45}{DEAL2}  &  \rotatebox[origin=c]{-45}{DEAL2-rpc}  &  \rotatebox[origin=c]{-45}{Oracle-2}\\\midrule

\multirow{3}{*}{\MultiLineCell{City/State} }
&  Head  &  \phantom{00 }\numprint{138} & \phantom{00 }\numprint{100} &  \PrintDecimal{2.71} &  \PrintDecimal{0.69} & \PrintDecimal{0.69} & \PrintDecimal{0.58} & \PrintDecimal{0.53} & \tb{\PrintDecimal{0.47}} & \PrintDecimal{0.39} & \PrintDecimal{2.65} & \PrintDecimal{0.77} & \tb{\PrintDecimal{0.55}} & \PrintDecimal{0.39} \\
&  Torso &   \phantom{0}\numprint{1746} &  \phantom{0}\numprint{1900} &  \PrintDecimal{5.13} &  \PrintDecimal{2.83} & \PrintDecimal{2.49} & \PrintDecimal{2.42} & \PrintDecimal{2.24} & \tb{\PrintDecimal{2.21}} & \PrintDecimal{1.62} & \PrintDecimal{4.08} & \PrintDecimal{2.02} & \tb{\PrintDecimal{1.96}} & \PrintDecimal{1.42} \\
&  Tail  &             \numprint{26946} &            \numprint{40793} & \PrintDecimal{18.65} & \PrintDecimal{15.40} & \PrintDecimal{13.45} & \PrintDecimal{13.42} & \tb{\PrintDecimal{13.41}} & \PrintDecimal{13.43} & \PrintDecimal{11.07} & \PrintDecimal{18.26} & \PrintDecimal{12.97} & \tb{\PrintDecimal{12.94}} & \PrintDecimal{10.59}\\\midrule

\multirow{ 3}{*}{City}
&  Head  &  \phantom{00 }\numprint{100} & --- & \PrintDecimal{3.94} & \PrintDecimal{2.12} & \PrintDecimal{2.31} & \PrintDecimal{2.28} & \PrintDecimal{2.02} & \tb{\PrintDecimal{1.79}} & \PrintDecimal{1.99} & \PrintDecimal{3.58} & \PrintDecimal{2.05} & \tb{\PrintDecimal{1.89}} & \PrintDecimal{1.69}\\
&  Torso &   \phantom{0}\numprint{1900} & --- & \PrintDecimal{9.50} & \PrintDecimal{7.02} & \PrintDecimal{7.06} & \PrintDecimal{7.01} & \PrintDecimal{6.20} & \tb{\PrintDecimal{5.91}} & \PrintDecimal{2.30} & \PrintDecimal{8.63} & \PrintDecimal{6.57} & \tb{\PrintDecimal{5.70}} & \PrintDecimal{1.87}\\
&  Tail  &             \numprint{25437} & --- & \PrintDecimal{29.70} & \PrintDecimal{22.84} & \tb{\PrintDecimal{22.69}} & \PrintDecimal{22.72} & \PrintDecimal{23.51} & \PrintDecimal{23.44} & \PrintDecimal{13.58} & \PrintDecimal{29.15} & \PrintDecimal{22.86} & \tb{\PrintDecimal{22.57}} & \PrintDecimal{13.12}\\\midrule

\multirow{ 3}{*}{\MultiLineCell{Music Title/Artist}}
&  Head  &  \phantom{00 }\numprint{178} & \phantom{00 }\numprint{100} &  \PrintDecimal{12.70} &  \PrintDecimal{9.97} & \PrintDecimal{9.20} & \PrintDecimal{9.04} & \tb{\PrintDecimal{8.78}} & \PrintDecimal{8.99} & \PrintDecimal{8.27} & \PrintDecimal{6.87} & \PrintDecimal{5.97} & \PrintDecimal{5.89} & \PrintDecimal{5.46}\\
&  Torso &   \phantom{0}\numprint{5405} &  \phantom{0}\numprint{3900} &  \PrintDecimal{17.48} & \PrintDecimal{14.53} & \tb{\PrintDecimal{12.79}} & \PrintDecimal{12.86} & \PrintDecimal{12.91} & \PrintDecimal{12.93} & \PrintDecimal{11.78} & \PrintDecimal{10.53} & \tb{\PrintDecimal{8.74}} & \PrintDecimal{8.79} & \PrintDecimal{8.03}\\
&  Tail  &             \numprint{82769} &            \numprint{60000} &  \PrintDecimal{57.70} & \PrintDecimal{50.19} & \tb{\PrintDecimal{45.01}} & \PrintDecimal{45.27} & \PrintDecimal{45.26} & \PrintDecimal{45.28} & \PrintDecimal{43.78} & \PrintDecimal{51.04} & \PrintDecimal{40.17} & \tb{\PrintDecimal{40.05}} & \PrintDecimal{38.50}\\\midrule

\multirow{ 3}{*}{\MultiLineCell{Music Title}}
&  Head  &  \phantom{00 }\numprint{100} & --- &  \PrintDecimal{11.42} &  \PrintDecimal{8.51} & \PrintDecimal{8.37} & \PrintDecimal{8.78} & \tb{\PrintDecimal{6.38}} & \PrintDecimal{7.00} & \PrintDecimal{3.02} & \PrintDecimal{10.53} & \PrintDecimal{7.49} & \PrintDecimal{6.68} & \PrintDecimal{2.67} \\
&  Torso &   \phantom{0}\numprint{3900} & --- &  \PrintDecimal{12.66} &  \PrintDecimal{9.97} & \PrintDecimal{9.90} & \PrintDecimal{9.57} & \PrintDecimal{9.59} & \tb{\PrintDecimal{9.15}} & \PrintDecimal{4.12} & \PrintDecimal{11.14} & \PrintDecimal{8.40} & \tb{\PrintDecimal{8.20}} & \PrintDecimal{3.25}\\
&  Tail  &             \numprint{60000} & --- &  \PrintDecimal{48.35} & \PrintDecimal{48.18} & \PrintDecimal{47.92} & \tb{\PrintDecimal{47.77}} & \PrintDecimal{48.21} & \PrintDecimal{47.81} & \PrintDecimal{34.42} & \PrintDecimal{45.89} & \tb{\PrintDecimal{45.48}} & \PrintDecimal{45.68} & \PrintDecimal{32.73}\\\midrule

\multirow{ 3}{*}{\MultiLineCell{Music Artist}}
&  Head  &  \phantom{00 }\numprint{100} & --- &  \PrintDecimal{11.37} &   \PrintDecimal{8.95} & \PrintDecimal{8.95} & \PrintDecimal{8.90} & \PrintDecimal{8.36} & \tb{\PrintDecimal{8.08}} & \PrintDecimal{5.34} & \PrintDecimal{9.24} & \PrintDecimal{7.23} & \tb{\PrintDecimal{6.13}} & \PrintDecimal{4.23}\\
&  Torso &  \phantom{00 }\numprint{900} & --- &  \PrintDecimal{14.30} &  \PrintDecimal{11.76} & \PrintDecimal{11.81} & \PrintDecimal{12.00} & \PrintDecimal{11.53} & \tb{\PrintDecimal{11.41}} & \PrintDecimal{8.65} & \PrintDecimal{11.27} & \PrintDecimal{9.80} & \tb{\PrintDecimal{9.45}} & \PrintDecimal{6.92}\\
&  Tail  &             \numprint{60000} & --- &  \PrintDecimal{59.59} & \PrintDecimal{100.00} & \PrintDecimal{57.74} & \tb{\PrintDecimal{57.37}} & \PrintDecimal{57.66} & \PrintDecimal{57.66} & \PrintDecimal{45.19} & \PrintDecimal{52.05} & \PrintDecimal{51.27} & \tb{\PrintDecimal{51.02}} & \PrintDecimal{40.04}\\\midrule

\multirow{ 3}{*}{\MultiLineCell{Soccer Athletes}}
&  Head  &  \phantom{00 }\numprint{100} & --- &  \PrintDecimal{6.25} & \tb{\PrintDecimal{5.44}} & \tb{\PrintDecimal{5.44}} & \tb{\PrintDecimal{5.44}} & \tb{\PrintDecimal{5.44}} & \tb{\PrintDecimal{5.44}} & \PrintDecimal{5.44} & \PrintDecimal{4.84} & \PrintDecimal{3.83} & \PrintDecimal{3.83} & \PrintDecimal{3.83}\\
&  Torso &  \phantom{00 }\numprint{300} & --- &  \PrintDecimal{17.12} & \PrintDecimal{12.65} & \PrintDecimal{12.65} & \PrintDecimal{12.72} & \tb{\PrintDecimal{12.57}} & \tb{\PrintDecimal{12.57}} & \PrintDecimal{11.71} & \PrintDecimal{15.10} & \PrintDecimal{10.48} & \tb{\PrintDecimal{10.41}} & \PrintDecimal{9.68}\\
&  Tail  &             \numprint{60000} & --- &  \PrintDecimal{78.63} & \PrintDecimal{62.90} & \PrintDecimal{62.95} & \tb{\PrintDecimal{62.88}} & \PrintDecimal{63.16} & \PrintDecimal{63.08} & \PrintDecimal{62.48} & \PrintDecimal{78.43} & \tb{\PrintDecimal{62.74}} & \PrintDecimal{62.85} & \PrintDecimal{62.42}\\

\midrule
{\MultiLineCell{General VA}}  &  --  &  -- & -- &  \tb{\PrintDecimal{12.04}} & \PrintDecimal{12.16} & \PrintDecimal{12.11} & \PrintDecimal{12.11} & \PrintDecimal{12.09} & \PrintDecimal{12.14} & \PrintDecimal{5.90} & \tb{\PrintDecimal{12.07}} & \PrintDecimal{12.11} & \PrintDecimal{12.11} & \PrintDecimal{5.93}\\
\bottomrule%
\end{tabular}%
}%
\end{table*}

\subsection{Data}
\label{data}
We tested our model using US English data. Our KG covered 5 entity types: city, state, music title, music artist, and soccer athlete.  Cities and states were connected with \textit{is in} and \textit{contains} relations.  Music titles and music artists were connected with \textit{performed by} and \textit{performed} relations. Our KG provided usage-based popularity estimates for each entity name and relationship as well as word counts for every entity name.  We chose these entity types and relations for our experiments to cover a variety of domains, popularity distributions, and entity name word counts.

Our training and testing data consisted of a combination of anonymized general VA data and synthesized data covering the targeted entities and relations.  We synthesized data to avoid the large expense of recording and manually transcribing data with adequate entity/relation coverage. From the general VA data, we used a manually transcribed uniform sample for training and testing.

To derive the templates and frequency estimates described in \autoref{features}, we started from a larger pool of automatically transcribed and named-entity-tagged VA requests. We first replaced regions tagged as named entities in each utterance with non-terminals, then extracted candidate templates sorted by count.  We then hand-curated the template lists, removing erroneous templates and cutting off the template lists when the templates became obscure.  Through inspection, we determined that 100 templates was a reasonable threshold for most entity types and entity type pairs.  The exceptions were soccer athletes and cities where only 10 and 79 templates were chosen, respectively.

To synthesize targeted data, as well as to support popularity-conditioned non-terminals, we divided the entities of each entity type into three strata, head, torso, and tail, according to their popularity estimates.  We determined the thresholds for the strata by inspecting the popularity distributions.  All of the distributions were L-shaped, declining steeply at the start and flattening at the end, with a gradual curve in the middle.  For all entity types and entity type pairs, the steep decline ended at about the 100th entity.  This determined the threshold for the head.  To set the lower bound on the torso, we chose the point at which the popularity distribution appeared to flatten out. We then included up to 60,000 entities in the tail.  The number of entities in each stratum for each entity type and entity type pair is provided in \autoref{tab:results}.  To synthesize targeted data for each stratum, we generated text data probabilistically according to the template frequency estimates and entity popularities in the KG. After generating text utterances, we synthesized audio for each text utterance using our previous generation speech synthesizer, a unit selection system described in~\cite{capes2017siri}.

To train the DEAL models, we used 7000 utterances from the general VA data and 2000 synthesized utterances for every entity type or entity type pair and stratum. To allay concerns that our approach would only work on the voice used for synthesizing our training data, we used a female voice to synthesize the training data and a male voice to synthesize test data.
\vspace{-4pt}
\subsection{Baseline Systems}
\vspace{-3pt}
Our ASR system used a deep convolutional neural network acoustic model \cite{Huang2020sndcnn}, 
a 4-gram LM with Good-Turing smoothing in the first pass, and the same LM interpolated with a Feed-Forward Neural Network (FFNN) LM \cite{Zhang2015fofe} in the second pass. For training and testing, we used the n-bests lists and lattices resulting from the second pass.  We trained our models starting from two baseline systems, baseline-1 and baseline-2, which differed in their 4-gram LMs. %

To build the 4-gram LM for baseline-1, we started by building component 4-gram models from data sources consisting of more than 10B manually and automatically transcribed anonymized VA requests.  We then combined these components using count merging to minimize perplexity on held out manually transcribed VA data \cite{bacchiani2003unsupervised}.  In baseline-2, we followed the same process but added LM components trained on synthetic text data.  We generated the synthetic text in a way similar to how we generated the synthesized training data for our model but in much larger quantities. We generated 100 million utterances for each entity type and entity type pair and trained a separate 4-gram LM component on each data set.  (This approach is similar to one of the approaches described in \cite{Gandhe2018sds}, where domain-specific LM components are built based on grammars and interpolated with more general components.)

\subsection{Results}

\begin{table}[t]%
\renewcommand{\arraystretch}{0.80}%
\setlength{\tabcolsep}{5pt}%
\small%
\centering%
\caption{Overview of variations being compared.\label{tab:methods}}%
\resizebox{0.42\textwidth}{!}{%
\begin{tabular}{ll l}%
\toprule
    Variation & Description & \# features  \\
    \midrule
    \tb{DEAL} & \EntityTypeEntityRelationship{} & \phantom{0 }\numprint{513} \\
    \tb{DEAL-r} & \tb{DEAL} + \EntityEntityRelationship{} & \phantom{0 }\numprint{560} \\
    \tb{DEAL-rc} & \tb{DEAL-r} + entity name word count & \numprint{2278} \\
    \tb{DEAL-rp} & \tb{DEAL-r} + entity popularity & \numprint{2278} \\
    \tb{DEAL-rpc} & \tb{DEAL-rp} $\bigcup$ \tb{DEAL-rc} & \numprint{6535} \\
    \bottomrule
\end{tabular}%
}%
\end{table}%
\begin{table}[t]
\renewcommand{\arraystretch}{0.80}%
\setlength{\tabcolsep}{5pt}%
\small%
\centering%
\caption{Examples of improvements on synthetic utterances resulting from DEAL.\label{tab:case_studies}}%
\resizebox{0.47\textwidth}{!}{%
    \begin{tabular}{ll}
    \toprule
        Baseline-2 & DEAL2-rpc\\
        \midrule
        directions to Hammers Texas &  directions to Amherst Texas\\
        turn on the traffic & Toronto traffic\\
        play Cinderella by young Jesus & play Cinderella by Yung Zeus \\
        lyrics tablet on my jeans & lyrics to Blood on my jeans \\
        play Brett Eldridge & play Brett Eldredge \\
        cadence on cavity age & Edinson Cavani age\\
    \bottomrule
    \end{tabular}}%
\end{table}%
\autoref{tab:results} show our experimental results in terms of Sentence Error Rate (SER) on test data using the model variations outlined in \autoref{tab:methods}.
We show results for all variations on top of baseline-1 (DEAL1) and selected variations on top of baseline-2 (DEAL2). We use SER (as opposed to WER) to measure performance, since getting an entity name only partially correct will often result in an incorrect VA response. We also include the oracle SER, which indicates the maximum possible improvement that can be achieved by rescoring the baseline lattices. We can see that SER of Oracle-2 is generally lower than Oracle-1, resulting from the inclusion of targeted synthetic data in the baseline LM. We also observe an increase in SER as we move from head to tail and torso across all models, since less popular entities will be less well represented in the general VA data used to train the baseline LMs.

We see that using DEAL1 results in significant improvement over baseline-1 on the synthetic datasets with minimal degradation on general VA queries. Comparing DEAL1-r and DEAL1, we see additional gains using DEAL1-r, mainly on test sets containing related entity pairs (Music Title/Artist and City/State).

We see improvements over DEAL1-r by adding word count (DEAL1-rc) and popularity (DEAL1-rp) features. Across most test sets DEAL1-rp  outperforms DEAL1-r and DEAL1-rc.  Combining both popularity and word count features (DEAL1-rpc) results in lower SER than either feature set on its own.

Before looking at the results for baseline-2 model variations, it is interesting to compare the performance of DEAL1-rpc and baseline-2.  We can see that DEAL1-rpc outperforms baseline-2 across several datasets, despite lacking the additional hypotheses available in baseline-2 lattices.

Finally, comparing DEAL2 and baseline-2, we notice that our method can achieve SER reductions even over a stronger baseline. We see further improvements by adding the \EntityEntityRelationship{} and including both word count and popularity features (DEAL2-rpc). \autoref{tab:case_studies} shows utterances for which DEAL2-rpc was able to correct the error made by the baseline-2 model.

%% file: 05-conclusions.tex
\section{Conclusions}
\label{ssec:conclusion}

We introduced a discriminative LM that incorporates real-world knowledge from a KG via n-gram features and a method for efficiently applying the model with lattice rescoring.  We evaluated the impact of using features that captured entity-entity relations, as well as features that took into account entity popularity and entity name word count. Our approach achieved relative SER reductions of more than 25\% on some synthetic test sets covering less common entities while causing minimal degradation on a general VA test set.  Including features that captured entity-entity relations led to large SER reductions on synthesized test sets covering related entity pairs, while including features conditioned on popularity and word count resulted in a modest SER reduction over all synthesized entity test sets.